\pdfoutput=1

\documentclass[runningheads]{llncs}
\usepackage{graphicx}
\usepackage{amsmath,amssymb} 
\usepackage{color}

\usepackage{float}
\floatstyle{plaintop}
\restylefloat{table}

\definecolor{mygreen}{RGB}{4, 202, 15}

\usepackage{comment}
\usepackage{algpseudocode,algorithm,algorithmicx}
\usepackage{caption}
\usepackage{subcaption}

\newcommand\samethanks[1][\value{footnote}]{\footnotemark[#1]}

\begin{document}
\pagestyle{headings}
\mainmatter

\title{Look Before You Leap: \\Bridging Model-Free and Model-Based Reinforcement Learning for Planned-Ahead Vision-and-Language Navigation} 

\titlerunning{Look Before You Leap}

\authorrunning{X. Wang et al.}

\author{Xin Wang\thanks{Equal contribution}, Wenhan Xiong\samethanks, Hongmin Wang, William Yang Wang}

\institute{
	University of California, Santa Barbara \\
	\email{ \{xwang,xwhan,hongmin\_wang,william\}@cs.ucsb.edu}
}

\maketitle

\begin{abstract}
Existing research studies on vision and language grounding for robot navigation focus on improving model-free deep reinforcement learning (DRL) models in synthetic environments. However, model-free DRL models do not consider the dynamics in the real-world environments, and they often fail to generalize to new scenes. In this paper, we take a radical approach to bridge the gap between synthetic studies and real-world practices---We propose a novel, planned-ahead hybrid reinforcement learning model that combines model-free and model-based reinforcement learning to solve a real-world vision-language navigation task. Our look-ahead module tightly integrates a look-ahead policy model with an environment model that predicts the next state and the reward. 
Experimental results suggest that our proposed method significantly outperforms the baselines and achieves the best on the real-world Room-to-Room dataset. Moreover, our scalable method is more generalizable when transferring to unseen environments.

\keywords{Vision-and-Language Navigation, First-Person View Video, Model-based Reinforcement Learning}
\end{abstract}

\section{Introduction}
It is rather trivial for a human to follow the instruction \textit{``Walk beside the outside doors and behind the chairs across the room. Turn right and walk up the stairs...''}, but teaching robots to navigate with such instructions is a very challenging task. The complexities arise from not just the linguistic variations of instructions, but also the noisy visual signals from the real-world environments that have rich dynamics.
Robot navigation via visual and language grounding is also a fundamental goal in computer vision and artificial intelligence, and it is beneficial for many practical applications as well, such as in-home robots, hazard removal, and personal assistants.

\begin{figure}[t]
\centering
\includegraphics[width=0.9\textwidth]{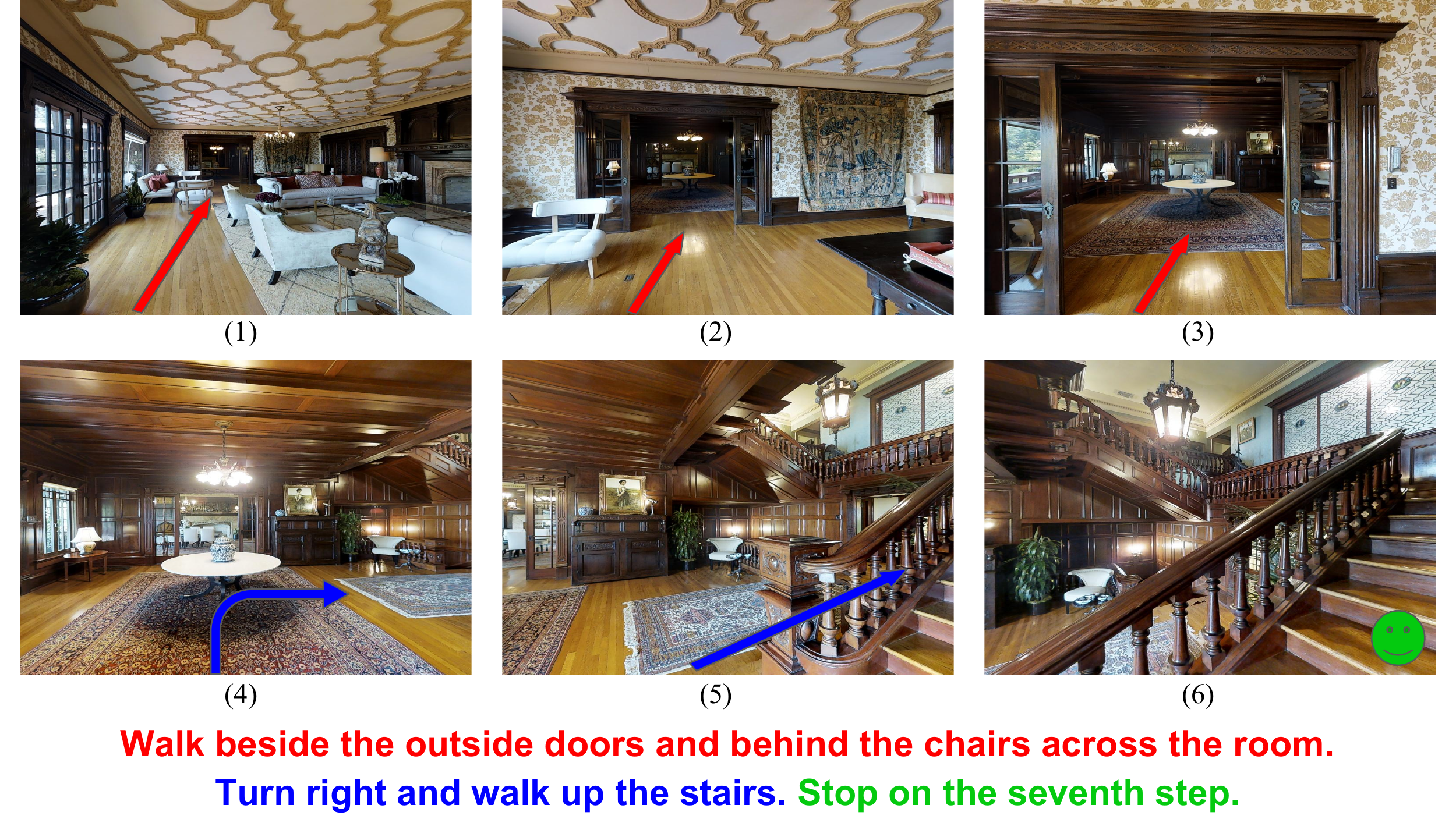}
\caption{An example of our task. The embodied agent learns to navigate through the room and arrive at the destination (\textbf{{\color{mygreen} green}}) by following the natural language instructions.  \textbf{{\color{red} Red}} and \textbf{{\color{blue} blue}} arrows match the orientations depicted in the pictures to the corresponding sentence. 
}
\label{fig:example}
\end{figure}

Vision-and-Language Navigation (VLN) is the task of training an embodied agent which has the first-person view as humans to carry out natural language instructions in the real world~\cite{anderson2017vision}. 
Figure~\ref{fig:example} demonstrates an example of the VLN task, where the agent moves towards to the destination by analyzing the visual scene and following the natural language instructions.
This is different from some other vision \& language tasks where the visual perception and natural language input are usually fixed (\textit{e.g.} Visual Question Answering). For VLN, the agent can interact with the real-world environment, and the pixels it perceives are changing as it moves. Thus, the agent must learn to map its visual input to the correct action based on its perception of the world and its understanding of the natural language instruction. 
 
Although steady progress has been made on the natural language command of robots~\cite{beattie2016deepmind,kempka2016vizdoom,zhu2017target,misra2017mapping}, it is still far from perfect. Previous methods are mainly employing \textit{model-free} reinforcement learning (RL) to train the intelligent agent by directly mapping raw observations into actions or state-action values. But model-free RL does not consider the environment dynamics and usually requires a large amount of training data. Besides, most of them are evaluated only in synthetic rather than real-world environments, which significantly simplifies the noisy visual \& linguistic perception problem, and the subsequent reasoning process in the real world.  

It is worth noticing that when humans follow the instructions, however, they do not solely rely on the current visual perception, but also imagine what the environment would look like and plan ahead in mind before actually performing a series of actions. For example, in baseball, the catcher and the outfield players often predict the direction and the rate of speed that the ball will travel, so they can plan ahead and move to the expected destination of the ball. Inspired by this fact, we seek the help of recent advance of \textit{model-based} RL~\cite{oh2017value,weber2017imagination} for this task. Model-based RL attempts to learn a model that can be used to simulate the environment and do multi-step lookaheads for planning. With an internal environment model to predict the future and plan ahead, the agent can benefit from the planning while avoiding from some trial-and-error in the real environment. 

Therefore, in this paper, we propose a novel approach which improves the vision-and-language navigation task performance by Reinforced Planning Ahead (which we refer as RPA). More specifically, our method, for the first time, endows the intelligent VLN agent with an environment model to simulate the world and predict the future visual perception. Thus the agent can realize directly mapping from the current real observation and planning of the future observations at the same time, and then perform an action based on both. Furthermore, We choose the real-world Room-to-Room (R2R) dataset as the testbed of our method. Our model-free RL model significantly outperforms the baseline methods as reported in the R2R dataset. Moreover, being equipped with the look-ahead module, our RPA model further improves the results and achieves the best on the R2R dataset. Hence, our contributions are three-fold:
\begin{itemize}
\item We are the first to combine model-free and model-based DRL for vision-and-language navigation.
\item Our proposed RPA model significantly outperforms the baselines and achieves the best on the real-world R2R dataset.
\item Our method is more scalable, and its strong generalizability allows it to be better transferred to unseen environments than the model-free RL methods.
\end{itemize}

\section{Related Work}
\subsubsection{Vision, Language and Navigation}
Recently, the intersection of vision and language research has attracted a lot of attention. Much work~\cite{xu2015show,vinyals2015show,karpathy2015deep,chen2015mind,yu2016video,wang2018watch,wang2018video,AREL2018} has been done in language generation conditioned on visual inputs. There is also another line of work~\cite{huang2016visual,antol2015vqa} that tries to answer questions from images. The task of vision-language grounding~\cite{thomason2017guiding,alomari2017learning,alomari2017natural} is more relevant to our task, which requires the ability to connect the language semantics to the physical properties of the environment. Our task requires the same ability but is more task-driven. The agent in our task needs to sequentially interact with the environment and finish a navigation task specified by a language instruction. 

Early approaches~\cite{kim1999symbolic,borenstein1989real,borenstein1991vector,oriolo1995line} on robot navigation usually require a prior global map or needs to build an environment map on-the-fly. The navigation goal in these methods is usually directly annotated in the map. In contrast to these work, the VLN task is more challenging in the sense that no global map is required and the goal is not directly annotated but described by natural language. Under this setting, several methods have been proposed recently. Mei \emph{et al.}~\cite{mei2016listen} proposed a sequence-to-sequence model to map the language to navigation actions. Misra \emph{et al.}~\cite{misra2017mapping} formulate navigation as a sequential-decision process and propose to use reward shaping to effectively train the RL agent. In the same environment, Xiong \emph{et al.}~\cite{xiong2018scheduled} propose a scheduled training mechanism which yields more efficient exploration and achieves better results. However, these methods still operate in synthetic environments and consider either simple discrete observation inputs or unrealistic top-down view of the environment.

\subsubsection{Model-based Reinforcement Learning} Using model-based RL for planning is a long-standing problem in reinforcement learning. Recently, the great computational power of neural networks makes it more realistic to learn a neural model to simulate environments~\cite{watter2015embed,lenz2015deepmpc,finn2017deep}. 
But for more complicated environments where the simulator is not exposed to the agent, the model-based RL usually suffers from the mismatch between the learned and real environments~\cite{gu2016continuous,talvitie2015agnostic}. 
In order to combat this issue, RL researchers are actively working on combining model-free and model-based RL~\cite{sutton1990integrated,yao2009multi,tamar2016value,silver2016predictron}. 
Most recently, Oh \textit{et al.}~\cite{oh2017value} propose a Value Prediction Network whose abstract states are trained to make predictions of future values rather than of future observations, and Weber \textit{et al.}~\cite{weber2017imagination} introduce an imagination-augmented agent to construct implicit plans and interpret predictions. 
Our algorithm shares the same spirit and is derived from these methods. But instead of testing on games, we, for the first time, adapt the combination of model-based and model-free RL for the real-world vision-and-language task. Another related work by Pathak \textit{et al.}~\cite{pathak2017curiosity} also learns to predict the next state during roll-out. An intrinsic reward is calculated based on the state prediction. Instead of inducing an extra reward, we directly incorporate the state prediction into the policy module. In other words, our agent takes into account the future predictions when making action decisions.

\section{Method}
\subsection{Task Definition}
As shown in Figure~\ref{fig:example}, we consider an embodied agent that learns to follow natural language instructions and navigate in realistic indoor environments. Specifically, given the agent's initial pose $p_0 = (v_0, \phi_0, \theta_0)$, which includes the spatial position, heading and elevation angles, and a natural language instruction (sequence of words) $\mathcal{X} = \{x_1,x_2, ..., x_n\}$, the agent is expected to choose a sequence of actions $\{a_1,a_2, ..., a_T\} \in \mathcal{A}$ and arrive at the target position $v_{target}$ specified by the language instruction $\mathcal{X}$. The action set $\mathcal{A}$ consists of six unique actions, \emph{i.e.} \textit{turn left, turn right, camera up, camera down, move forward}, and \textit{stop}. In order to figure out the desired action $a_t$ at each time step, the agent needs to effectively associate the language semantics with its visual observation $o_t$ about the environment. Here the observation $o_t$ is the raw RGB image captured by the mounted camera. The performance of the agent is evaluated by both the success rate $P_{succ}$ (the percentage of test instructions that are correctly followed by the agent) and the final navigation error $E_{nav}$ (average final distance from the target position).

\begin{figure}[t]
\centering
\includegraphics[width=1\textwidth]{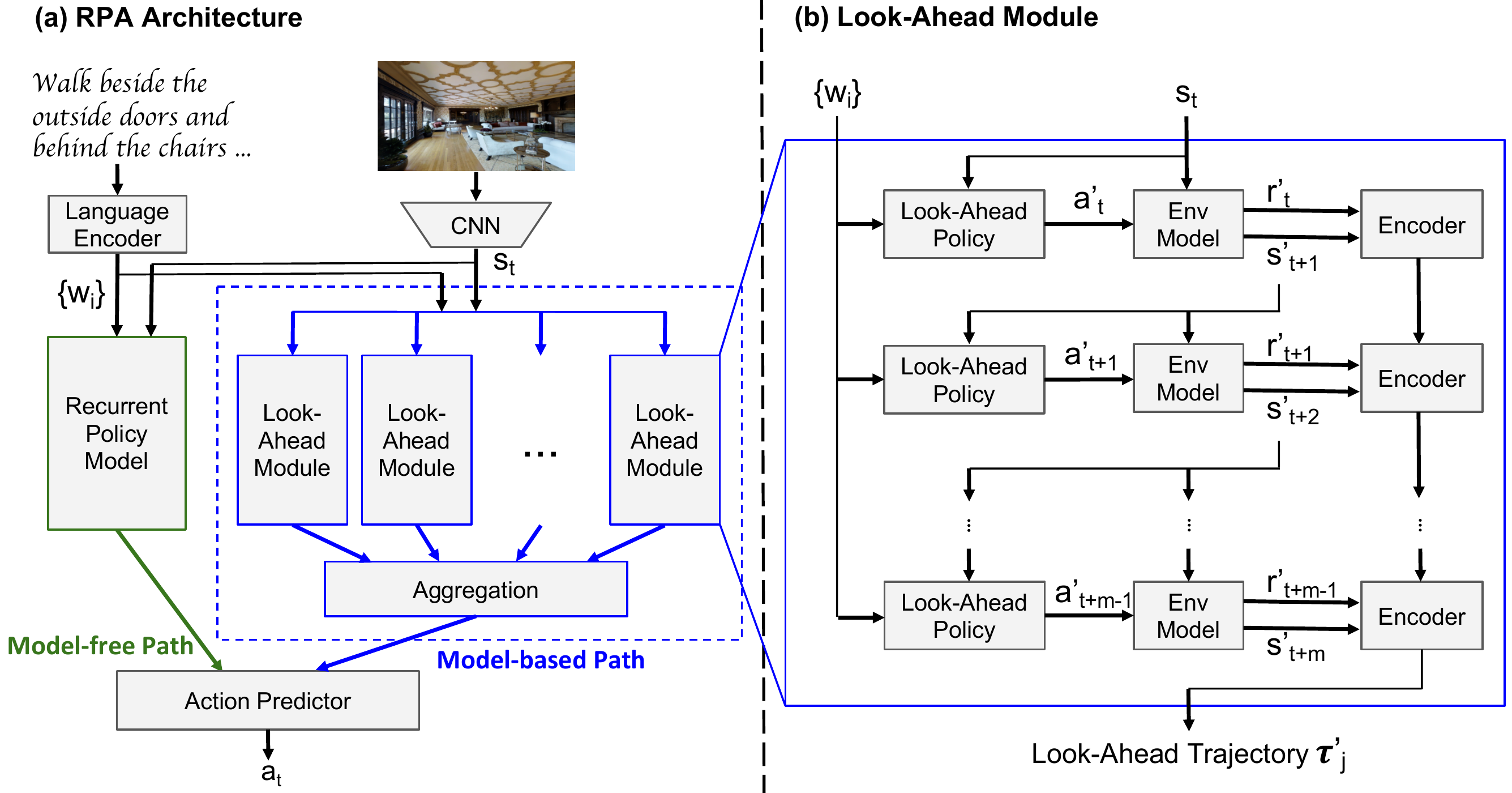}
\caption{The overview of our method.}
\label{fig:overview}
\end{figure} 

\subsection{Overview}
In consideration of the sequential-decision making nature of the VLN task, we formulate VLN as a reinforcement learning problem, where the agent sequentially interacts with the environments and learns by trial and error. Once an action is taken, the agent receives a scalar reward $r(a_t,s_t)$ from the environment. The agent's action $a_t$ at each step is determined by a parametrized policy function $\pi(o_t;\theta)$. The training objective is to find the optimal parameters $\theta$ that maximize the discounted cumulative rewards:
\begin{equation}
\max_{\theta} \mathcal{J^{\pi}} = \mathbb{E} \Big[ \sum_{t=1}^{T} \gamma^{t-1} r(a_t,s_t) | \pi(o_t;\theta) \Big] \quad,
\end{equation}
where $\gamma \in (0,1)$ is the discounted factor that reflects the significance of future rewards. 

We model the policy function as a sequence-to-sequence neural network that encodes both the language sequence $\mathcal{X} = \{x_1,x_2, ...,x_n\}$ and image frames $\mathcal{O} = \{o_1,o_2, ...,o_T\}$ and decodes the action sequence $\{a_1,a_2, ..., a_T\}$. The basic model consists of a \textbf{language encoder} that encodes the instruction $\mathcal{X}$ as word features $\{w_1,w_2, ...,w_n\}$, an \textbf{image encoder} that extracts high-level visual features,  and a \textbf{recurrent policy network} that decodes actions and recurrently updates its internal state, which is supposed to encode the history of previous actions and observations. To reinforce the agent by planning ahead and further improve the model's capability, we equip the agent with \textbf{look-ahead modules}, which employ the \textbf{environment model} to take into account the future predictions.

As illustrated in Figure~\ref{fig:overview}(a), at each time step $t$, the recurrent policy model takes as input the word features $\{w_i\}$ and the state $s_i$ and produces the information for the final decision making, which forms a \textit{model-free path} by itself. In addition, the \textit{model-based path} exploits multiple look-ahead modules to realize look-ahead planning and imagine the possible future trajectories. 
The final action $a_t$ is chosen by the \textbf{action predictor}, based on the information from both the model-free and model-based paths. Therefore, our RPA method seamlessly integrates model-free and model-based reinforcement learning.

\subsection{Look-Ahead Module}
The core component of the RPA method is the look-ahead module, which is used to imagine the consequences of planning ahead multiple steps from the current state $s_t$. In order to augment the agent with imagination, we introduce the \textit{environment model} that makes a prediction about the future based on the state of the present. 
Since directly predicting the raw RGB image $o_{t+1}$ is very challenging, our environment model, instead, attempts to predict the abstract-state representation $s_{t+1}$ that represents the high-level visual feature.

Figure~\ref{fig:overview}(b) showcases the internal process of the look-ahead module, which consists of an environment model, a look-ahead policy, and a trajectory encoder. Given the abstract-state representation $s_t$ of the real world at step $t$, the look-ahead policy\footnote{We adopt the recurrent policy used in the model-free path as the look-ahead policy in all our experiments.} first takes $s_t$ as input and outputs an imagined action $a'_t$. Our environment model receives the state $s_t$ and the action $a'_t$, and predicts the corresponding reward $r'_t$ and the next state $s'_{t+1}$. Then the look-ahead policy will take a further action $a'_{t+1}$ based on the predicted state $s'_{t+1}$. The environment model will make a new prediction $\{r'_{t+1}, s'_{t+2}\}$. This look-ahead planning goes $m$ steps, where $m$ is the preset trajectory length. We use an LSTM to encode all the predicted rewards and states along the look-ahead trajectory and outputs its representation $\tau'_{j}$. As shown in Figure~\ref{fig:overview}(a), at every time step $t$, our model-based path operates $J$ look-ahead processes and we obtain a look-ahead trajectory representation $\tau'_{j}$ for each ($j = 1,...,J$). These $J$ look-ahead trajectories are then aggregated (by concatenation) together and passed to the action predictor as the information of the model-based path.

\begin{figure}[t]
\centering
\includegraphics[width=6.5cm]{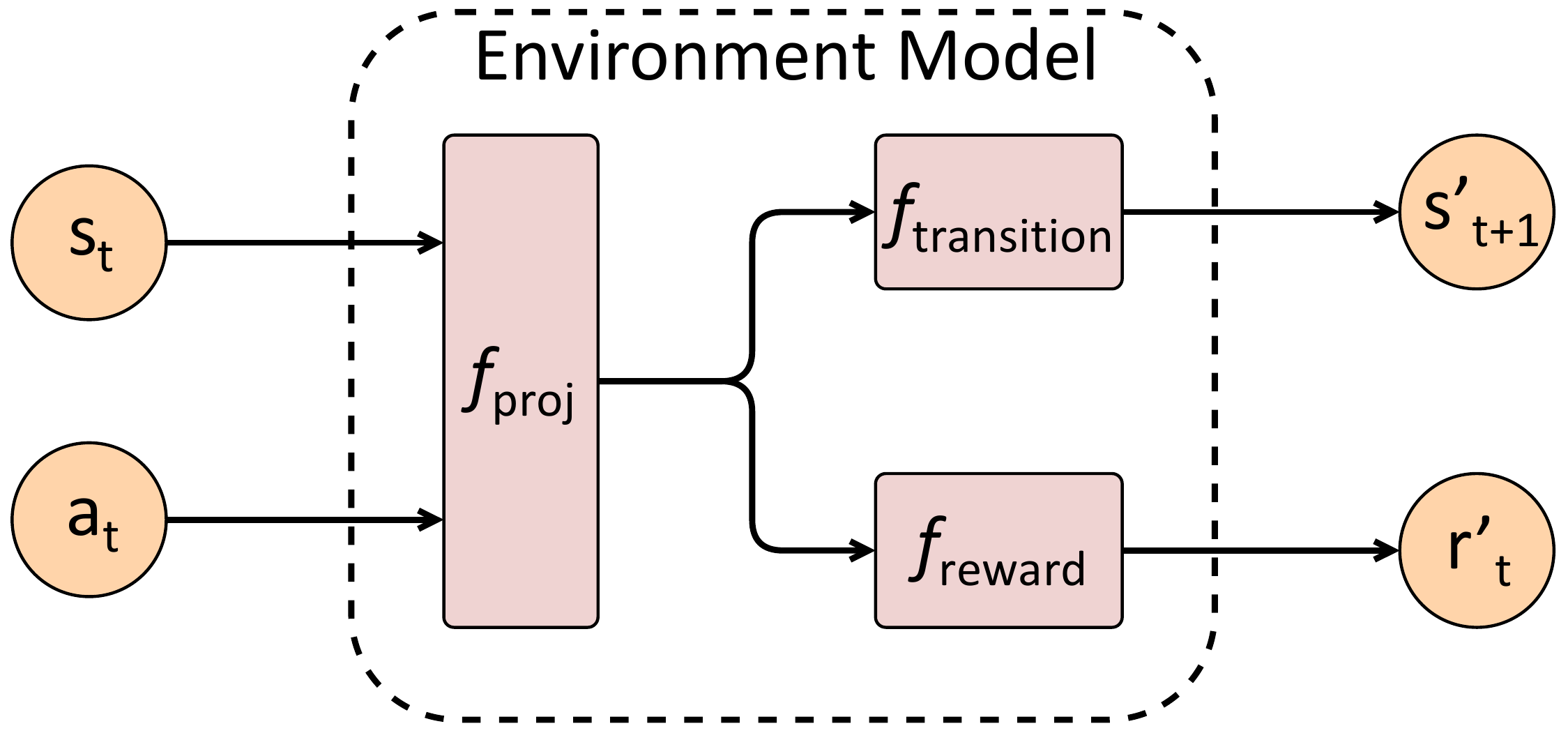}
\caption{The environment model.}
\label{fig:env}
\end{figure} 

\subsection{Models}
Here we further discuss the architecture designs of the learnable models in our methods that are not specified above, including the environment model, the recurrent policy model, and the action predictor. 


\subsubsection{Environment Model}
Given current state $s_t$ and the action $a_t$ taken by the agent, the environment model predicts the next state $s'_{t=1}$ and the reward $r'_t$. As is shown in Figure~\ref{fig:env}, the projection function $f_{proj}$ first concatenates $s_t$ and $a_t$ and then projects them into the same feature space. Its output is then fed into the transition function $f_{transition}$ and the reward function $f_{reward}$ to obtain $s'_{t=1}$ and $r'_t$ respectively. In formula,
\begin{align}
s'_{t+1} &= f_{transition}(f_{proj}(s_t, a_t)) \\
r'_t &= f_{reward}(f_{proj}(s_t, a_t)) \quad,
\end{align}
where $f_{proj}$, $f_{transition}$, and $f_{reward}$ are all learnable neural networks. Specifically, $f_{proj}$ is a linear projection layer, $f_{transition}$ is a multilayer perceptron with sigmoid output, and $f_{reward}$ is also a multilayer perceptron but directly outputs the scalar reward.     

\subsubsection{Recurrent Policy Model}

\begin{figure}[t]
\centering
\includegraphics[width=11cm]{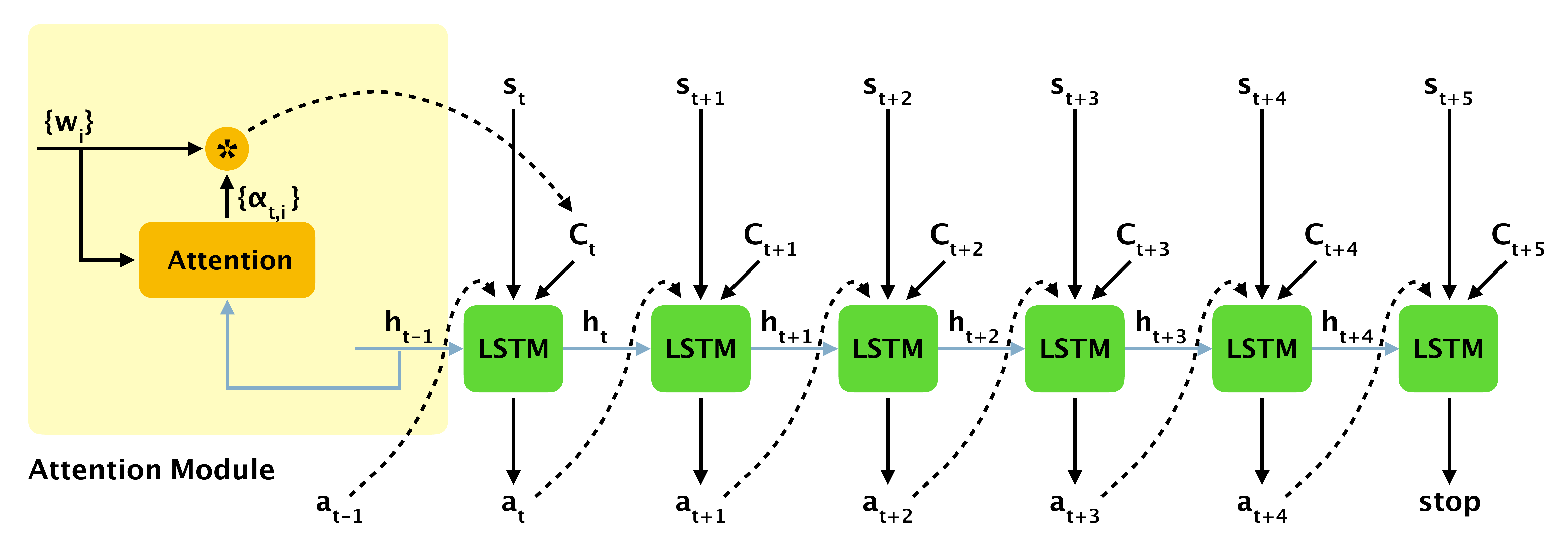}
\caption{An example of the unrolled recurrent policy model (from $t$ to $t+5$). The left-side yellow region demonstrates the attention mechanism at time step $t$.}
\label{fig:policy}
\end{figure}

Our recurrent policy model is an attention-based LSTM decoder network (see Figure~\ref{fig:policy}). At each time step $t$, the LSTM decoder produces the action $a_t$ by considering the context of the word features $\{w_i\}$, the environment state $s_t$, the previous action $a_{t-1}$, and its internal hidden state $h_{t-1}$. Note that one may directly take the encoded word features $\{w_i\}$ as the input of the LSTM decoder. We instead adopt an attention mechanism to better capture the dynamics in the language instruction and dynamically put more attention to the words that are beneficial for the current action selection.  

The left-hand side of Figure~\ref{fig:policy} is a demo attention module for the LSTM decoder. At each time step $t$, the context vector $c_t$ is computed as a weighted sum over the encoded word features $\{w_i\}$
\begin{equation}
c_t = \sum \alpha_{t,i} w_i \quad .
\end{equation}
These attention weights $\{\alpha_{t,i}\}$ act as an alignment mechanism by giving higher weights to certain words which match the decoder's current status, and are defined as
\begin{equation} \label{eq:att1}
    \alpha_{t,i} = \frac{\exp(e_{t,i})}{\sum_{k=1}^n \exp(e_{t,k})} \quad,
    \quad \text{where}~ e_{t,i} = h_{t-1}^\top w_i \quad.
\end{equation}
$h_{t-1}$ is the decoder's hidden state at previous step. 

Once the context vector $c_t$ is obtained, the concatenation of $[c_t, s_t, a_{t-1}]$ is fed as the input of the decoder to produce the intermediate model-free feature for the action predictor's use.
Formally, 
\begin{align}
h_t &= LSTM(h_{t-1}, [c_t, s_t, a_{t-1}]) \quad .
\end{align}
Then the output feature is the concatenation of the LSTM's output $h_t$ and the context vector $c_t$, which will be passed to the action predictor for making the decision. But if the recurrent policy model is employed as an individual policy (\textit{e.g.} the look-ahead policy), then it directly outputs the action $a_t$ based on $[h_t; c_t]$. Note that in our model, we feed the context vector $c_t$ to both the LSTM and the output posterior, which boosts the performance than solely feeding it into the input.

\subsubsection{Action Predictor}
The action predictor is a multilayer perceptron with a SoftMax layer as the last layer. Given the information from both the model-free and model-based paths as the input, the action predictor generates a probability distribution over the action space $\mathcal{A}$.  

\subsection{Learning}
The training of the whole system is a two-step process: learning the environment model first and then learning the enhanced policy model, which is equipped with the look-ahead module. It is worth noting that the environment model and policy model have their own language encoders and are trained separately. The environment model will be fixed during policy learning. 

\subsubsection{Environment Model Learning}
Ideally, the look-ahead module is expected to provide the agent with accurate predictions of future observations and rewards. If the environment model is noisy itself, it can actually provide misleading information and make the training even more unstable. In terms of this, before we plug in the look-ahead module, we pretrain the environment model using a randomized teacher policy. Under this policy, the agent will decide whether to take the human demonstration action or a random action based on a Bernoulli meta-policy with $p_{human} = 0.95$. Since the agent's policy will get closer to demonstration (optimal) policy during training, the environment model trained by demonstration policy will help it better predict the transitions close to the optimal trajectories. On the other hand, in reinforcement learning methods, the agent's policy is usually stochastic during training. Making the agent take the random action under the probability of $1 - p_{human}$ is to simulate the stochastic training process. We define two losses to optimize this environment model:
\begin{align}
\mathit{l}_{transition} &= \mathbb{E}[(s'_{t+1} - s_{t+1})^2] \\
\mathit{l}_{reward} &= \mathbb{E}[(r'_{t+1} - r_{t+1})^2] \quad .
\end{align}
The parameters are updated by jointly minimizing these two losses.

\subsubsection{Policy Learning}
With the pretrained environment model, we can incorporate the look-ahead module into the policy model. We first discuss the general pipeline of training the RL agent and then describe how to train the proposed RPA model.

In the VLN task, two distinct supervisions can be used to train the policy model. First, we can use the demonstration actions provided by the simulator to do pure supervised learning. The training objective in this case is to simply maximize the log-likelihood of the demonstration action:

\begin{equation}
\mathcal{J}_{sl} =  \mathbb{E} [ \log(\pi(a_h|o;\theta)) ] \quad ,
\end{equation}
where $a_h$ is the demonstration action. This agent can quickly learn a policy that perform relative well on seen scenes. However, pure supervised learning only encourage the agent to imitate the demonstration paths. This potentially limits the agent's ability to recover from erroneous actions in an unseen environment. To also encourage the agent to explore the state-action space outside the demonstration path, we utilize the second supervision, \emph{i.e.} the reward function. The reward function depends on the environment state $s$ and agent's action $a$, and is usually not differentiable in terms of $\theta$. As the objective of the VLN task is to successfully arrive at the target position, we define our reward function based on the distance metric. We denote the distance between a state $s$ and the target position $v_{target}$ as $\mathcal{D}_{target}(s)$. Then the reward after taking action $a_t$ at state $s_t$ is defined as:
\begin{equation} \label{i_r}
r(s_t,a_t) = \mathcal{D}_{target}(s_{t}) - \mathcal{D}_{target}(s_{t+1}) \quad .
\end{equation}
It indicates whether the action reduces the agents distance from the target. Obviously, this reward function only reflects the immediate effect of a particular action but ignores the action's future influence. To account for this, we reformulate the reward function in a discounted cumulative form:
\begin{equation}
\label{r}
R(s_t,a_t) =  \sum_{t'=t}^{T} \gamma^{t'-t}r(s_{t'},a_{t'})  \quad .
\end{equation}
Besides, the success of the whole trajectory can also be used as an additional binary reward. Further details on reward setting are discussed in the experiment section. With the reward function, the RL objective then becomes:
\begin{equation}
\mathcal{J}_{rl} = \mathbb{E}_{a \sim \pi(\theta)} [ \sum_t{R(s_t,a_t)} ] \quad .
\end{equation}
Using the likelihood-ratio estimator in the REINFORCE algorithm, the gradient of $\mathcal{J}_{rl}$ can be written as: 
\begin{equation}
\nabla_{\theta}\mathcal{J}_{rl} = \mathbb{E}_{a \sim \pi(\theta)} [\nabla_{\theta} \log \pi(a|s;\theta) R(s,a)] \quad .
\end{equation}
With this two training objective, we can either use a mixed loss function as in~\cite{ranzato2015sequence} to train the whole model, or use the supervised learning to warm-start the model and use RL to do fine-tuning. In our case, we find the mixed loss converges faster and achieves better performance.

\begin{algorithm}[t] 
    \caption{RL training with planning ahead}
    \label{alg:training}
    \begin{algorithmic}[1]
        \State $\theta_p$: policy parameters to be learned, $\theta_e$: environment model parameters
        \State Initialize the R2R environment
        \While{not converged}
            \State Roll-out a trajectory $(<s_1,a_1,r_1>,<s_2,a_2,r_3>,...,<s_T,a_t,r_T>)$ 
            \State Update $\theta_e$ using $ g  \propto \nabla_{\theta_e}(\mathit{l}_{transition} + \mathit{l}_{reward})$
           \EndWhile
        \For{iteration=0,M-1}
            \State initialize the weight for supervised loss $w_{SLloss} \leftarrow 1$
            \State Sample a batch of training instructions
            \State $s_0 \leftarrow $ initial state
            \For{$t$ = 0, MAX\_EPISODE\_LEN-1}
                
                   \State Perform depth-bounded ($depth=2$) roll-outs using the environment model
                \State Use roll-out encoder to encoder all these simulated
                \State Sample actions under the current policy in parallel
                \State Save immediate rewards $r(s_t,a_t)$ and performed actions $a_t$
                \If{All Ended}
                    \State Break
                \EndIf
            \EndFor
            \State Compute the discounted cumulative reward $R(s_t,a_t)$
            \State Total loss $\mathit{l}_{policy} = - w_{SLloss} * \mathcal{J}_{sl} - (1 - w_{SLloss}) * \mathcal{J}_{rl}$
            \State Decrease $w_{SLloss}$: $w_{SLloss}\leftarrow 0.1 + 0.9 * \exp(iteration / \mathcal{T})$
            \State Update $\theta_p$ using $g \propto \nabla \mathit{l}_{policy}$
        \EndFor
    \end{algorithmic}
\end{algorithm}

To joint train the policy model and look-ahead module, we first freeze the pretrained environment model. Then at each step, we perform simulated depth-bounded roll-outs using the environment model. Since we have five unique actions besides the \textit{stop} action, we perform the corresponding five roll-outs. Each path is first encoded using an LSTM. The last hidden states of all paths are concatenated and then feed into action predictor. Now the learnable parameters come from three components: the original model-free policy mode, the roll-out encoder, and the action predictor. The pseudo-code of the algorithm is shown in Algorithm~\ref{alg:training}.

\section{Experiments}
\subsection{Experimental Settings}
\subsubsection{R2R Dataset}
Room-to-Room (R2R) dataset~\cite{anderson2017vision} is the first dataset for vision-and-language navigation task in real 3D environments. The R2R dataset is built upon the Matterport3D dataset~\cite{chang2017matterport3d}, which consists of 10,800 panoramic views constructed from 194,400 RGB-D images of 90 building-scale scenes (Many of the scenes can be viewed in the Matterport 3D spaces gallery\footnote{\url{https://matterport.com/gallery/}}). The R2R dataset further samples 7,189 paths capturing most of the visual diversity in the dataset and collects 21,567 navigation instructions with an average length of 29 words (each path is paired with 3 different instructions). As reported in \cite{anderson2017vision}, the R2R dataset is split into training (14,025 instructions), seen validation (1,020), unseen validation (2,349), and test (4,173) sets. Both the unseen validation and test sets contain environments that are unseen in the training set, while the seen validation set shares the same environments with the training set.

\subsubsection{Implementation Details}
We develop our algorithms on the open source code of the Matterport3D simulator\footnote{\url{https://github.com/peteanderson80/Matterport3DSimulator}}. ResNet-152 CNN features~\cite{he2016deep} are extracted for all the images without fine-tuning. 
In the model-based path, we perform one look-ahead planning for each possible action in the environment. The $j$-th look-ahead planning corresponds to the $j$-th of the action set $\mathcal{A}$, and the subsequent actions are executed by the shared look-ahead policy. In our experiments, we use the same policy model trained in the model-free path as the look-ahead policy. All the other hyperparameters are tuned on the validation set. More training details can be found in the supplementary material.

\subsubsection{Evaluation Metrics}
Following the conventional wisdom, the R2R dataset mainly evaluates the results by three metrics: \textit{navigation error}, \textit{success rate}, and \textit{oracle success rate}. We also report the \textit{trajectory length} though it is not a metric. The navigation error is defined as the shortest path distance in the navigation graph between the agent's final position $v_T$ and the destination $v_{target}$. The success rate calculates the percentage of the result trajectories whose navigation errors are less than 3m. The oracle success rate is also reported: the distance between the closest point on the trajectory and the destination is used to calculate the error, even if the agent does not stop there.   

\subsubsection{Baselines}
In the R2R dataset, there exists a ground-truth shortest-path trajectory (\textit{Shortest}) for each instruction sequence from the starting location $v_0$ to the target location $v_{target}$. This shortest-path trajectory can be further used for supervised training. \textit{Teacher-forcing}~\cite{luong2015effective} uses cross-entropy loss to train the model at each time step to maximize the likelihood of the next gound-truth action given the previous ground-truth action. 
Instead of feeding the ground-truth action back to the recurrent model, one can sample an action based on the output probabilities over the action space (\textit{Student-forcing}).
In our experiments, we list the results of these two models as reported in \cite{anderson2017vision} as our baselines. We also include the results of a random agent (\textit{Random}), which randomly takes an action at each step. 

\begin{table}[t]
\setlength{\tabcolsep}{2pt}
\begin{center}
  \begin{tabular}{ l  | c c c c | c c c c | c c c c}
      &    \multicolumn{4}{c|}{\textbf{Val Seen}} & \multicolumn{4}{c|}{\textbf{Val Unseen}} & \multicolumn{4}{c}{\textbf{Test (unseen)}} \\
      \hline
       \textbf{Model}    
    & \begin{tabular}{@{}c@{}} TL \\ (m) \end{tabular} 
    & \begin{tabular}{@{}c@{}} NE \\ (m) \end{tabular} 
    & \begin{tabular}{@{}c@{}} SR \\ (\%) \end{tabular} 
    & \begin{tabular}{@{}c@{}} OSR \\ (\%) \end{tabular} 
    & \begin{tabular}{@{}c@{}} TL \\ (m) \end{tabular} 
    & \begin{tabular}{@{}c@{}} NE \\ (m) \end{tabular} 
    & \begin{tabular}{@{}c@{}} SR \\ (\%) \end{tabular} 
    & \begin{tabular}{@{}c@{}} OSR \\ (\%) \end{tabular} 
    & \begin{tabular}{@{}c@{}} TL \\ (m) \end{tabular} 
    & \begin{tabular}{@{}c@{}} NE \\ (m) \end{tabular} 
    & \begin{tabular}{@{}c@{}} SR \\ (\%) \end{tabular} 
    & \begin{tabular}{@{}c@{}} OSR \\ (\%) \end{tabular} \\
      
      \hline
        Shortest 
        & 10.19 & 0.00 & 100 & 100 
        &9.48 & 0.00 & 100 & 100  
        & 9.93 & 0.00 & 100 & 100 \\
        Random 
        & 9.58 & 9.45 & 15.9 & 21.4
        & 9.77 & 9.23 & 16.3 & 22.0  
        & 9.93 & 9.77 & 13.2 & 18.3 \\
        Teacher-forcing
        & 10.95 & 8.01 & 27.1 & 36.7 
        & 10.67 & 8.61 & 19.6 & 29.1
        & - & - & - & -\\
        Student-forcing
        & 11.33 & 6.01 & 38.6 & 52.9
        & 8.39 & 7.81 & 21.8 & 28.4
        & 8.13 & 7.85 & 20.4 & 26.6 \\
    
    \hline
    \textbf{Ours} & \multicolumn{4}{c|}{} & \multicolumn{4}{c|}{} & \\
    XE 
    & 11.51 & 5.79 & 40.2 & \textbf{54.1} 
    & 8.94 & 7.97 & 21.3 & 28.7 
    & 9.37 & 7.82 & 22.1 & 30.1\\
    Model-free RL 
    & 10.88 & 5.82 & 41.9 & 53.5 
    & 8.75 & 7.88 & 21.5 & 28.9 
    & 8.83 & 7.76 & 23.1 & 30.2\\
    RPA 
    & 8.46 & \textbf{5.56} & \textbf{42.9} & 52.6 
    & 7.22 &\textbf{7.65} & \textbf{24.6} & \textbf{31.8}
    & 9.15 & \textbf{7.53} & \textbf{25.3} & \textbf{32.5}\\
  \end{tabular}
\end{center}
  \caption{Results on both the validation sets and test set in terms of four metrics: Trajectory Length (TL), Navigation Error (NE), Success Rate (SR), and Oracle Success Rate (OSR). We list the best results as reported in \cite{anderson2017vision}, of which Student-forcing performs the best. Our RPA method significantly outperforms the previous best results, and it is also noticeable that we gain a larger improvement on the unseen sets, which proves that our RPA method is more generalized.}
\label{table:result}
\vspace*{-3ex}
\end{table}

\subsection{Results and Analysis}
Table~\ref{table:result} shows the result comparison between our models and the baseline models. We first implement our own recurrent policy model trained with the cross-entropy loss (\textit{XE}). Note that our XE model performs better than the Student-forcing model on the test set. By switching to the model-free RL, the results are slightly improved. Then our RPA learning method further boosts the performance consistently on the metrics and achieves the best results in the R2R dataset, which validates the effectiveness of combining model-free and model-based RL for the VLN task.

An important fact revealed here is that our RPA method brings a notable improvement on the unseen sets and the improvement is even larger than that on the seen set (the relative success rates are improved by 6.7\% on Val Seen, 15.5\% on Val Unseen, and 14.5\% on Test over XE). While the model-free RL method gains a very small performance boost on the unseen sets. This proves our claim that it is easy to collect and utilize data in a scalable way to incorporate the look-ahead module for the decision making. Besides, our RPA method turns out to be more generalized and can be better transferred to unseen environments.  

\subsection{Ablation Study}

\subsubsection{Learning Curves of the Environment Model}
\begin{figure}[t]
\centering
\begin{subfigure}[t]{0.45\textwidth}
\centering
\includegraphics[height=3.5cm]{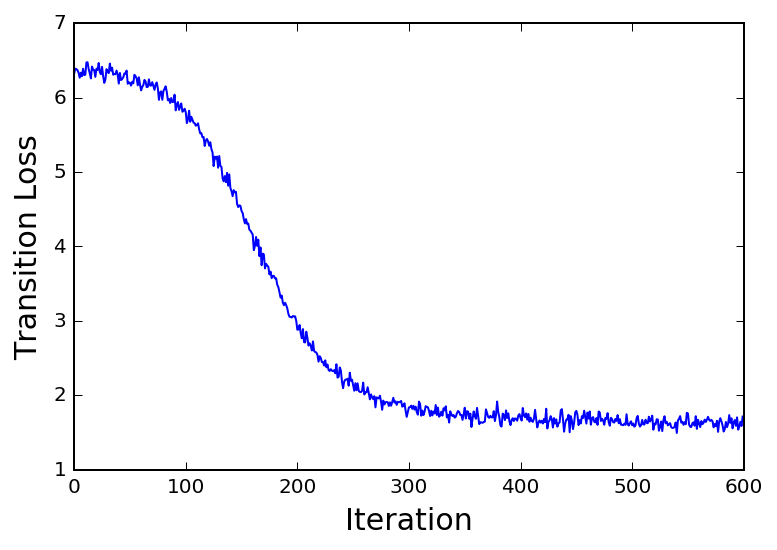}
\end{subfigure}
~
\begin{subfigure}[t]{0.45\textwidth}
\centering
\includegraphics[height=3.5cm]{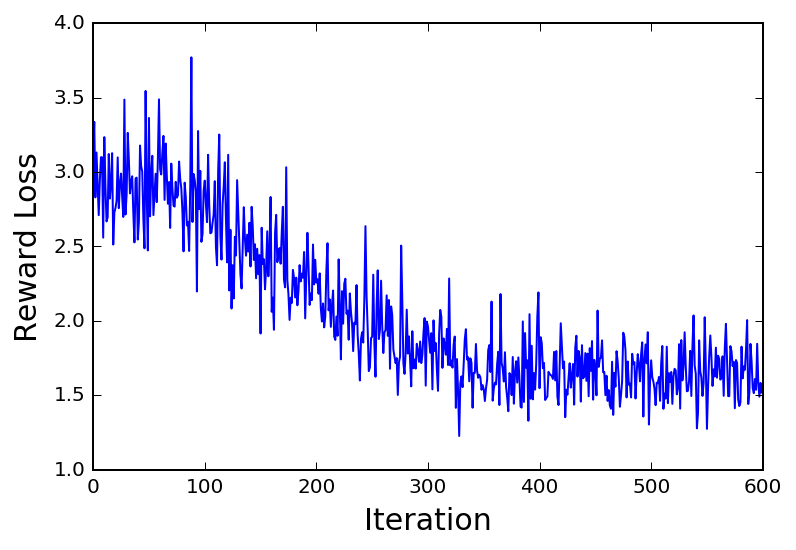}
\end{subfigure}
\caption{Learning curves of the environment model.}
\label{fig:env_loss}
\end{figure}

To realize our RPA method, we first need to train an environment model to predict the future state given the present state, which would be then plugged into the look-ahead module. So it is important to guarantee the effectiveness of the pretrained environment model. In Figure~\ref{fig:env_loss}, we plot both the transition loss and the reward loss of the environment model during training. Evidently, both losses converge to a stable point after around 500 iterations. 
But it is also noticeable that the learning curve of the reward loss is much noisier than that of the transition loss. This is because of the sparsity nature of rewards. Unlike the state transitions that are usually more continuous, the rewards within trajectory samples are very sparse and of high variance, thus it is noisier to predict the exact reward using mean square error.    

\subsubsection{Effect of Different Rewards}
We test four different reward functions in our experiments. The results are shown in Table~\ref{table:reward}. The \textit{Global Distance} reward function is defined per path by assigning the same reward to all actions along this path. This reward measures how far the agent approaches the target by finishing the path. The \textit{Success} reward is a binary reward: if the path is correct, then all actions will be assigned with a reward $1$, otherwise reward $0$. The \textit{Discounted} reward is defined as in Equation~\ref{r}. Finally, the \textit{Discounted} \& \textit{Success} reward, which is used by our final model, basically adds the \textit{Success} binary reward to the immediate reward (see Equation~\ref{i_r}) of the final action. Then the discounted cumulative reward is calculated using the Equation~\ref{r}. In the experiments, the first two rewards are much less effective than the discounted reward functions which assign different rewards to different actions. We believe the discounted reward calculated at every time step can better reflect the true value of each action. As the final evaluation is not only based on the navigation error but also success rate, we also observe that incorporating the success information into the reward can further boost the performance in terms of success rate.
\begin{table}[t]
\small
\setlength{\tabcolsep}{3pt}
\begin{center}
  \begin{tabular}{ l  | c c c | c c c }
      &    \multicolumn{3}{c|}{\textbf{Val Seen}} & \multicolumn{3}{c}{\textbf{Val Unseen}} \\
      \hline
       \textbf{Reward}    
    & \begin{tabular}{@{}c@{}} Navigation \\ Error \\ (m) \end{tabular} 
    & \begin{tabular}{@{}c@{}} Success \\ (\%) \end{tabular} 
    & \begin{tabular}{@{}c@{}} Oracle \\ Success \\ (\%) \end{tabular} 
    & \begin{tabular}{@{}c@{}} Navigation \\ Error \\ (m) \end{tabular} 
    & \begin{tabular}{@{}c@{}} Success \\ (\%) \end{tabular} 
    & \begin{tabular}{@{}c@{}} Oracle \\ Success \\ (\%)\end{tabular} \\
      \hline
        \textit{Global Distance} & 6.17 & 35.5 & 45.1 & 8.20 & 19.0 & 25.6  \\
        \textit{Success} & 6.21 & 37.8 & 43.2 & 8.17 & 21.3 & 26.7 \\
        \textit{Discounted} & \textbf{5.79} & 40.5 & 52.8 & \textbf{7.74} & 20.4 & 28.5   \\

        \textit{Discounted} $\&$ \textit{Success} & 5.82 & \textbf{41.9} & \textbf{53.5} & 7.88 & \textbf{21.5} & \textbf{28.9} \\
  \end{tabular}
\end{center}
\caption{Results of the model-free RL with different reward definitions.}
\label{table:reward}
\vspace*{-3ex}
\end{table} 

\subsubsection{Case Study}
\begin{figure}[t]
\centering
\includegraphics[width=0.7\textwidth]{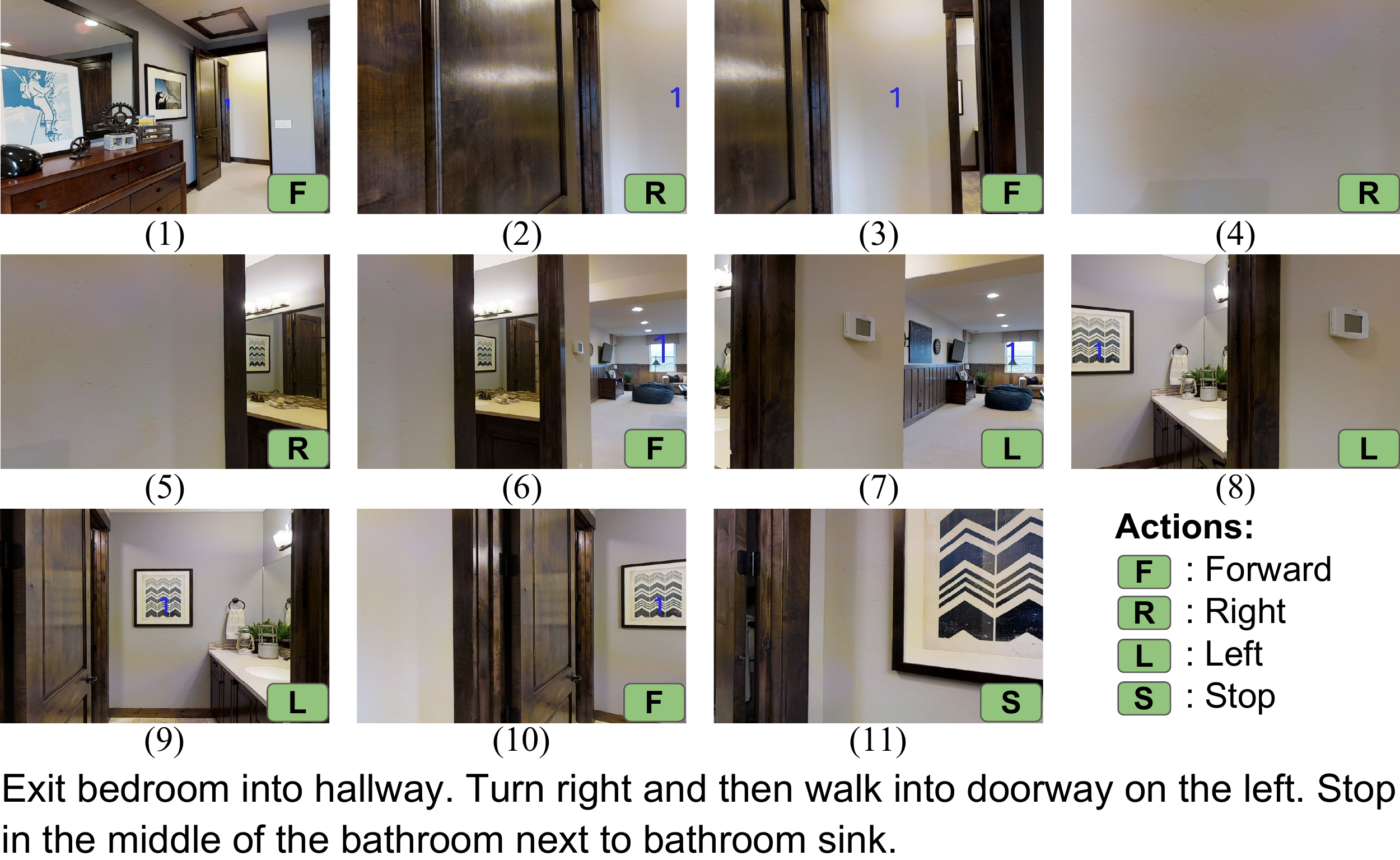}
\caption{An example trajectory executed by our RPA agent. Given the instruction and the starting position (1), the agent produces one action per time step. In this example we show all the 11 steps of this trajectory.}
\label{fig:case_study}
\vspace*{-2ex}
\end{figure}

For a more intuitive view of the decision-making process in the VLN task, we show a test trajectory that is performed by our RPA agent in Figure~\ref{fig:case_study}. The agent starts from position (1) and takes a sequence of actions by following the natural language instruction until it reaches the destination (11) and stops there. We observe that although the actions include \textit{Forward, Left, Right, Up, Down,} and \textit{Stop}, the action \textit{Up} and \textit{Down} appear very rare in the result trajectories. In most cases, the agent can still reach the destination even without moving up/down the camera, which indicates that the R2R dataset has its limitation on the action distribution.  

\section{Conclusion}
Through experiments, we demonstrate the superior performance of our proposed RPA approach, which also tackles the common generalization issue of the model-free RL when applying to unseen scenes. Besides, equipped with the look-ahead module, our method can simulate the environment and incorporate the imagined trajectories, making the model more scalable than the model-free agents. 
In the future, we plan to explore the potential of the model-based RL to transfer across different tasks, \textit{i.e.} Vision-and-Language Navigation, Embodied Question Answering~\cite{embodiedqa} etc.

\bibliographystyle{splncs04}
\bibliography{egbib}

\newpage
\appendix


\section{Error Analysis}
\label{sec:error}

\begin{figure}[t]
\centering
\begin{subfigure}[t]{1\textwidth}
  \centering
  \includegraphics[width=0.8\textwidth]{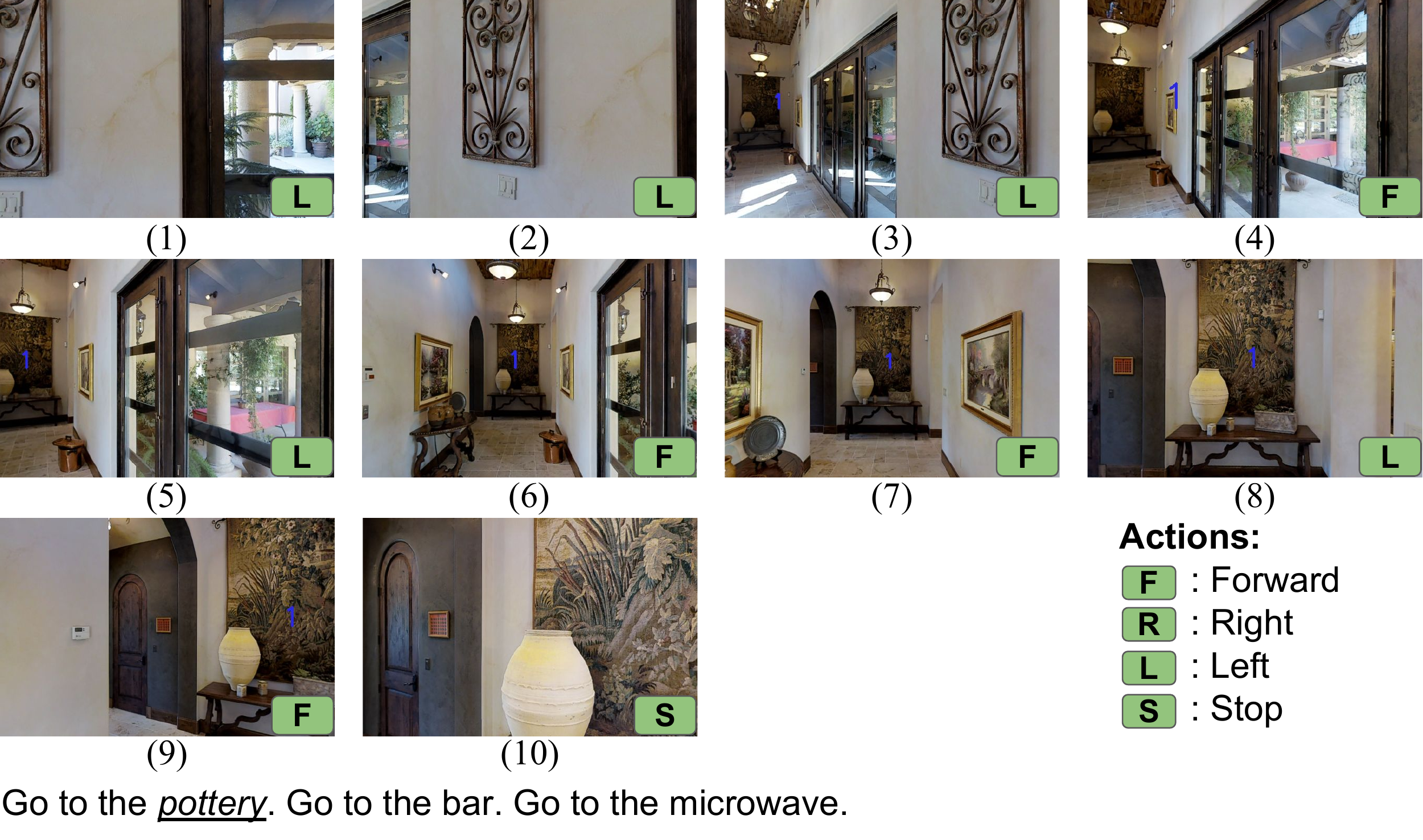}
  \vspace{-1ex}
  \caption{An error case with the OOV word \textit{pottery}.}
  \label{fig:oov}
\end{subfigure}
~\\\vspace{1ex}
\begin{subfigure}[t]{1\textwidth}
  \centering
  \includegraphics[width=0.8\textwidth]{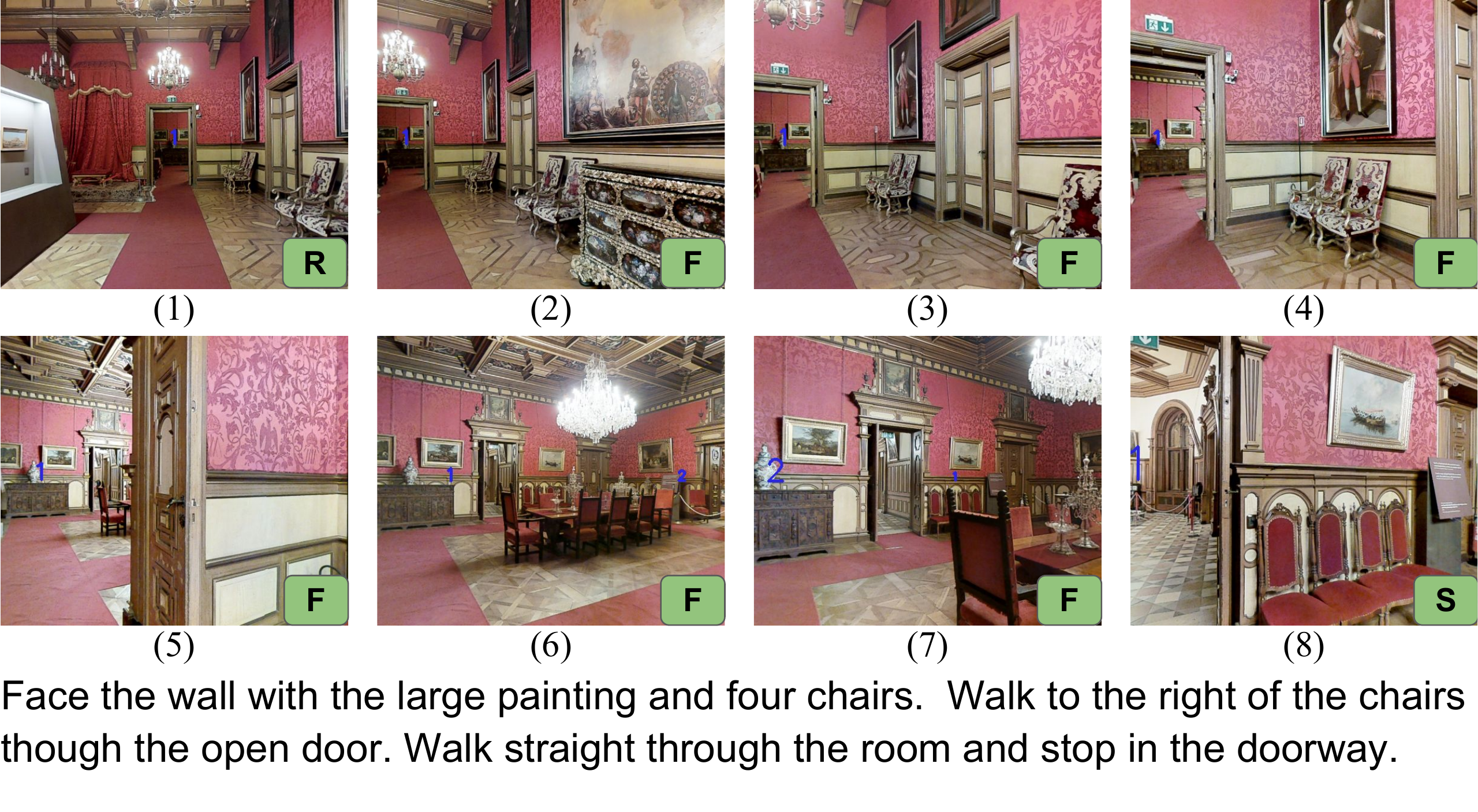}
  \vspace{-1ex}
  \caption{Complex unseen environment and ambiguous instruction.}
  \label{fig:unseen}
\end{subfigure}
~\\\vspace{1ex}
\begin{subfigure}[t]{1\textwidth}
  \centering
  \includegraphics[width=0.8\textwidth]{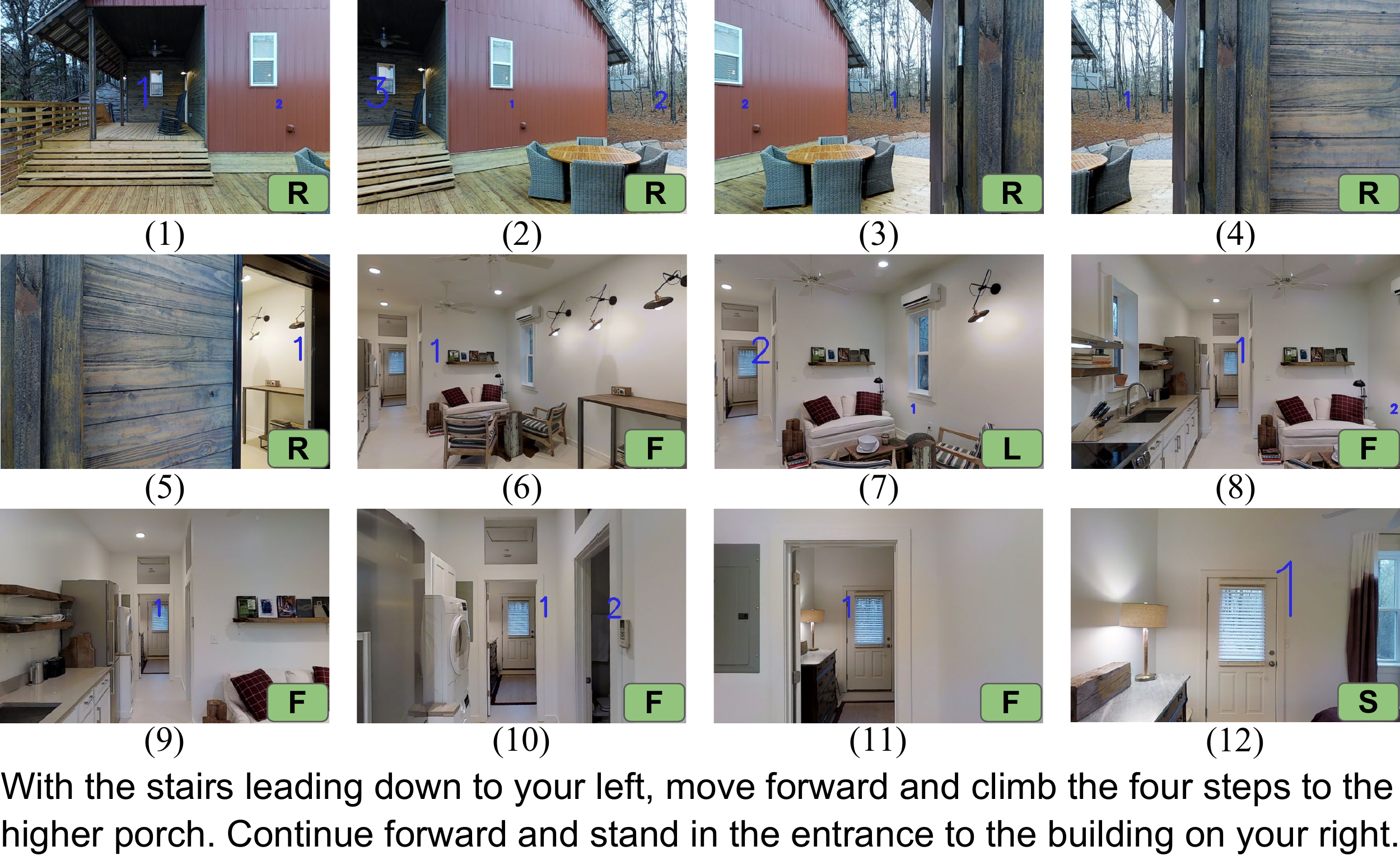}
  \vspace{-1ex}
  \caption{Error accumulation.}
  \label{fig:accumulation}
\end{subfigure}
\vspace*{-1ex}
\caption{Error cases. Please see Section~\ref{sec:error} for explanation.}
\label{fig:error}
\end{figure}

In addition to the quantitative results on the evaluation metrics, here we further analyze the negative results and demonstrate three kinds of common errors in the vision-and-language navigation task. 

First, the existing agents lack the ability to understand the instructions with out-of-vocabulary (OOV) words. During training, the agent can only see the words in the vocabulary and there are no special mechanisms aiming to resolve the OOV issue. Figure~\ref{fig:error}(a) illustrates such an error case where an OOV word (\textit{pottery}) appears in the instruction, and the agent fails to identify the \textit{pottery} on its right in the scene and mistakenly turns left and then goes forward. Utilizing external knowledge might be a good way to relieve the OOV issue. 

Moreover, we show in Figure~\ref{fig:error}(b) another error case where the agent is required to follow an ambiguous instruction in a relatively complex unseen environment. More specifically, it is asked to ``\textit{face the wall with the large painting and four chairs}'', but there are paintings on every wall in the scene, how \textit{large} the painting should be? The instruction actually refers to the larger one shown in the top right corner of the picture~(2), but the agent fails to perceive it and thus performs the wrong actions. Therefore it is a must for the agent to have a better understanding of the instruction and the ability to reason about the visual scene to avoid such errors.

The last case we discuss here is the error accumulation issue. Once the agent chooses some wrong actions, it is very likely that the sight of the agent is completely changed. So if the agent cannot fix the errors by itself, the instruction is not ``correct'' anymore based on the new scene after the wrong actions are taken. Figure~\ref{fig:error}(c) demonstrates one of the error accumulation cases where one bad decision leads to a series of bad decisions during the navigation process. All these issues mentioned above remain to be solved in future work.

\section{Network Architecture}
In this section, we provide the details of the neural network architectures used for the experiments. 

\subsubsection{Language Encoder}
The language encoder is a long short-term memory (LSTM) network with hidden size 512. It takes as input the sequence of word embeddings of the natural language instruction. The word embedding dimension is 256. Then the outputs of the LSTM are passed through a linear layer (512,512) to obtain the final output. Therefore, it follows: Input $\rightarrow$ Embedding (256) $\rightarrow$ LSTM (512) $\rightarrow$ Linear (512, 512) $\rightarrow$ Tanh $\rightarrow$ Output.

\subsubsection{Recurrent Policy Model}
The recurrent policy model consists of an action embedding layer of size 32, an LSTM decoder with hidden size 512, a dot-product attention module, and a projection module (Linear (1024, 512) $\rightarrow$ Tanh $\rightarrow$ Linear (512, 6) $\rightarrow$ SoftMax) that projects the concatenation of the decoder's output and the context vector into the probabilities of all actions. Note that if the recurrent policy model works as an individual policy, then it outputs the probabilities of all actions, which is the output of the projection module; while when employed in the model-free path, it directly passes the concatenation of the decoder's output and the context vector as the representation of the model-free path (the projection module is not used).

\subsubsection{Environment Model}
As shown in Figure~\ref{fig:env}, the environment model is composed of a projection function (Linear (256+2048, 512) $\rightarrow$ ReLU), transition function (Linear (512, 256) $\rightarrow$ ReLU $\rightarrow$ Linear (256, 512) $\rightarrow$ ReLU $\rightarrow$ Linear (512, 2048) $\rightarrow$ Sigmoid), and a reward function (Linear (512, 256) $\rightarrow$ ReLU $\rightarrow$ Linear (256, 1)). Besides, the environment model has its own action embedding layer of size 256.

\subsubsection{Trajectory Encoder}
The trajectory encoder is an LSTM encoder of size 256, which encodes the sequence of the predicted states and rewards (by concatenation). Its output, the encoded vector, is the hidden state of the LSTM at the last step $h_t$. We concatenate the encoded vectors of all look-ahead planning as the final representation of the model-based path.

\subsubsection{Action Predictor}
The action predictor is a multilayer perceptron: Linear (256$\times$5+1024, 512) $\rightarrow$ ReLU $\rightarrow$ Linear (512, 256) $\rightarrow$ ReLU $\rightarrow$ Linear (256, 6) $\rightarrow$ SoftMax.

\section{Training Details}
We tune all the hyperparameters on the validation set. Here we show the hyperparameter settings of both the RPA agent training and the environment model training.
\subsubsection{RPA Agent Hyperparameter Setting}
We set the maximal length of the action path as 20. The batch size is 100 and the maximum number of iterations is 40,000. We use the Adam optimizer with a learning rate of 1e-4 to optimize all parameters. To avoid exploding gradients, we clip the gradients of all parameters with a norm of 5. We also use an L2 weight decay of $0.0005$ and a dropout ratio of 0.5 to prevent overfitting. The discounted factor of our cumulative reward is 0.95. For the mixed loss, we initialize the weight of the supervised loss as 1.0 and set its lower bound as 0.15. In other words, the weight of supervised loss will never be less than 0.15. We observe that smaller weights of the supervised loss often lead to worse performance on the test samples in seen environments. At every time step, we run 5 individual look-ahead planning, each corresponds to one of the valid actions (except \textit{STOP}). The look-ahead length that achieves the best performance is 2.

\subsubsection{Environment Model Hyperparameter Setting}
The batch size is also set as 100. We use the Adam optimizer with a learning rate of 1e-5. We also use a L2 weight decay of $0.0005$ and a dropout ratio of 0.5 to prevent overfitting. The final loss is a weighted sum of the transition loss and the reward loss, whose weights are 1 and $0.001$ correspondingly. We notice that both the transition loss and reward loss converge to a stable point after around 500 iterations.

\end{document}